%% file: main.tex
\documentclass[10pt,twocolumn,letterpaper]{article}

\usepackage[pagenumbers]{iccv} 


\definecolor{iccvblue}{rgb}{0.21,0.49,0.74}
\usepackage[pagebackref,breaklinks,colorlinks,allcolors=iccvblue]{hyperref}
\usepackage{times}
\usepackage{soul}
\usepackage{booktabs}
\usepackage{algorithm}
\usepackage{algorithmic}
\usepackage[switch]{lineno}
\usepackage{multirow}
\usepackage{amsmath}

\def\paperID{12967} 
\def\confName{ICCV}
\def\confYear{2025}

\title{Extremely Simple Out-of-distribution Detection for Audio-visual Generalized Zero-shot Learning}

\author{
Yang~Liu$^{1}$, Xun~Zhang$^{1}$, Jiale~Du$^{1}$, Xinbo~Gao$^{1,2}$, Jungong~Han$^{3}$ \\
$^{1}$Xidian University, Xi'an, China \\
$^{2}$Chongqing University of Posts and Telecommunications, Chongqing, China \\
$^{3}$Tsinghua University, Beijing, China \\
{\tt\small yangl@xidian.edu.cn, xunz724@gmail.com, 23011211070@stu.xidian.edu.cn, }\\{\tt\small xbgao@mail.xidian.edu.cn, jungonghan77@gmail.com}
}

\begin{document}
\maketitle
\input{sec/0_abstract}    
\input{sec/1_intro}

\input{sec/2_related}

\input{sec/3_method}
\input{sec/4_experiment}

\input{sec/5_conclusion}

{
    \small
    \bibliographystyle{ieeenat_fullname}
    \bibliography{main}
}

\end{document}

%% file: sec/0_abstract.tex
\begin{abstract}
Zero-shot Learning~(ZSL) attains knowledge transfer from seen classes to unseen classes by exploring auxiliary category information, which is a promising yet difficult research topic. In this field, Audio-Visual Generalized Zero-Shot Learning~(AV-GZSL) has aroused researchers' great interest in which intricate relations within triple modalities~(audio, video, and natural language) render this task quite challenging but highly research-worthy. However, both existing embedding-based and generative-based AV-GZSL methods tend to suffer from domain shift problem a lot and we propose an extremely simple Out-of-distribution~(OOD) detection based AV-GZSL method~(EZ-AVOOD) to further mitigate bias problem by differentiating seen and unseen samples at the initial beginning. EZ-AVOOD accomplishes effective seen-unseen separation by exploiting the intrinsic discriminative information held in class-specific logits and class-agnostic feature subspace without training an extra OOD detector network. Followed by seen-unseen binary classification, we employ two expert models to classify seen samples and unseen samples separately. Compared to existing state-of-the-art methods, our model achieves superior ZSL and GZSL performances on three audio-visual datasets and becomes the new SOTA, which comprehensively demonstrates the effectiveness of the proposed EZ-AVOOD.
\end{abstract}

%% file: sec/1_intro.tex
\section{Introduction} 
\label{sec:intro}
Multimodal learning has become one of the most trending research topics nowadays involving vision-language tasks including visual question answering~\cite{vqa}, image captioning~\cite{captioning}, cross-modal retrieval~\cite{crossmodal}, visual entailment~\cite{entailment}, and so on, and audio-language tasks like audio captioning~\cite{audiocaptioning}, emotion recognition~\cite{emotion}, speech translation~\cite{translation}, etc., since vision, audio, and language are the most common signal sources in real world. However, it is undoubtedly an overwhelming burden to collect and label quantities of videos/pictures, audio signals, and natural language corpus to fulfill aforementioned tasks, where Zero-shot Learning~(ZSL) emerges as an attainable approach to get rid of redundant data collection by mining auxiliary information like semantic attributes and word embeddings to achieve knowledge transfer without contact with unseen samples. Concerning video-audio multimodal tasks, AV-GZSL dependent on the fusion of audio and visual input, and natural language description is the chosen methodology to cope with classification, retrieval, and other problems.

\begin{figure}[t]
    \centering
    \includegraphics[width=0.9\linewidth]{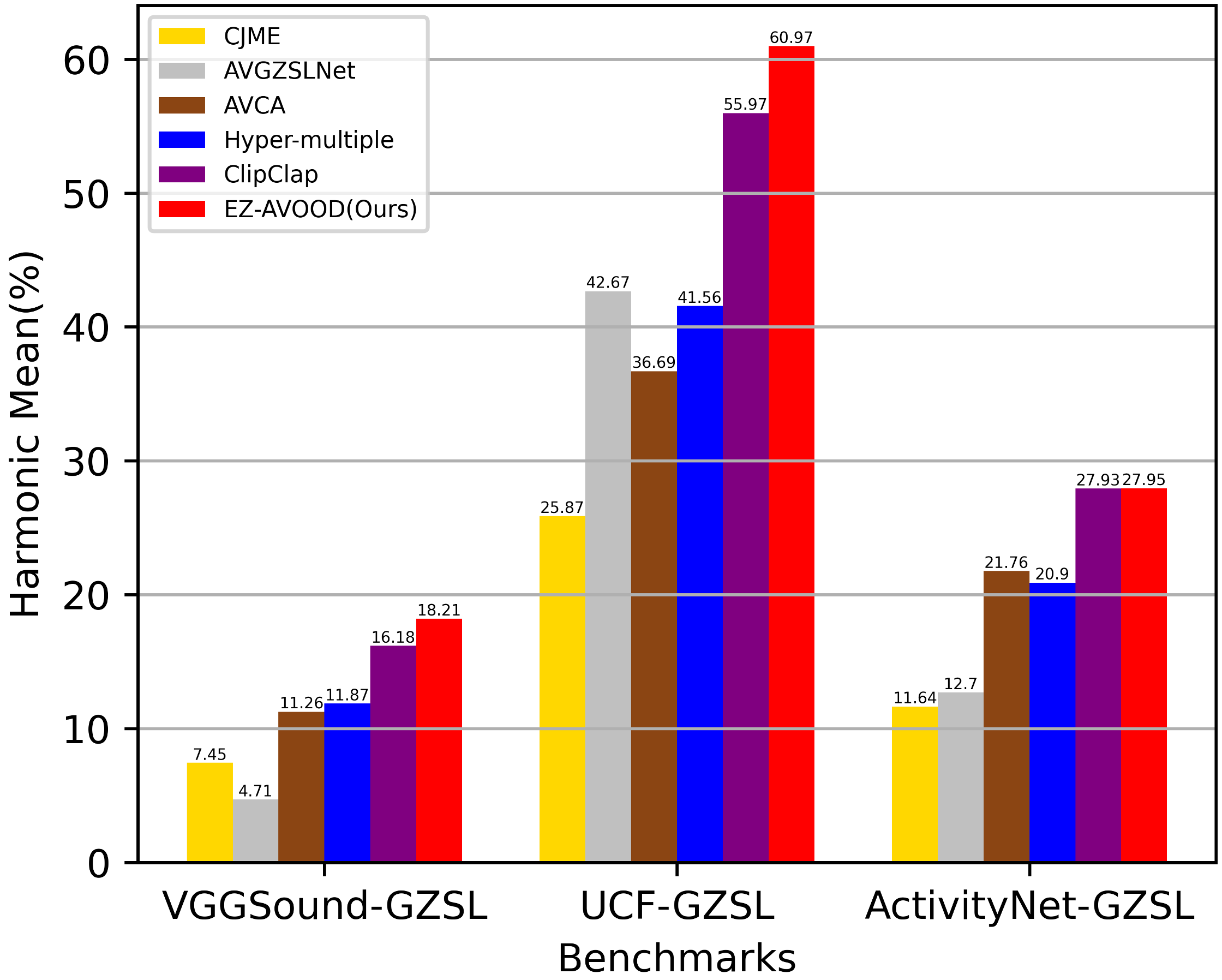}
    \caption{Harmonic mean~(\%) evaluating GZSL performance of our EZ-AVOOD model and other comparison methods on three datasets. EZ-AVOOD~(the red bar) consistently outperforms the rest opponents with a lead margin up to 5\% on the UCF-GZSL benchmark.}
    \label{front}
\end{figure}

\cite{cjme} sets out to apply GZSL to audio-visual features. AVGZSLNet~\cite{AVGZSLNet} leverages late fusion for integrating information from bi-modalities. Most of the subsequent works have focused on efficient fusion of audio-visual feature representations: AVCA~\cite{avca} initially utilizes cross-attention~\cite{attention} block which is inherited by \cite{temporal,hyperbolic,ezavgzl}. Additionally, generative approach AVFS~\cite{avfs} facilitates contrastive learning with the synthesized negative unseen samples. However, domain shift problem is always unavoidable for ZSL with the bias towards seen classes, AV-GZSL simply behaves the same way, especially for embedding-based approaches. Generative solutions aim to reduce the bias with synthesized unseen samples and Calibrated Stacking~\cite{calibrated} searches extra hyper-parameters to suppress the model's tendency towards seen samples. Considering that generative model is unstable to train while Calibrated Stacking's effect is limited, in this paper we propose an extremely simple OOD detection based AV-GZSL method named EZ-AVOOD to effectively alleviate the bias problem in an explicit way. Figure \ref{front} fully exhibits the competence of our model where our model outperforms all the contrasting methods on three audio-visual benchmarks.  

Typically, the fundamental framework of OOD approaches in ZSL shares the universal three parts: an OOD detector, one seen classifier and an unseen classifier, and each of them plays a vital role in the final GZSL performance. Additionally, these three components indeed work quite independently and consequently they are relatively substitutable when stronger components are developed. Unlike previous work which utilizes generative method like WGAN-GP~\cite{wgan,wgan-gp} to first synthesize unseen samples features, and then jointly train a completely new OOD detector with seen samples and synthesized unseen samples~\cite{avood}, the proposed EZ-AVOOD model accomplishes OOD detection without training the OOD detector with the aid of class-specific logits produced by supervised seen classifier and class-agnostic information hidden in feature subspace. More specifically, our method only needs to train the seen classifier composed of one MLP~(Multilayer Perceptrons), and fine-tune an existing embedding-based method \cite{clipclap} as the unseen expert classifier, which considerably reduces the complexity of the entire model. 
Moreover, the unseen classifier of our EZ-AVOOD can be substituted by arbitrary optimal AV-GZSL method for future researchers to achieve higher overall GZSL performance. 
The main contributions of this paper can be summarized into the following three aspects:
\begin{itemize}
  \item We proposed an extremely simple OOD detection based model EZ-AVOOD to address AV-GZSL problem with OOD score derived from class-specific logits and class-agnostic feature subspace instead of training a completely new OOD detector. 
  \item We comprehensively demonstrate the effectiveness of EZ-AVOOD model through practical ZSL and GZSL experiments on three different audio-visual benchmarks and observe a substantial enhancement in comparison with the current state-of-the-art methodologies. 
  \item The proposed effective OOD detection method EZ-OOD possesses strong compatibility with existing AV-GZSL approaches, which indicates that future researchers could effortlessly improve the GZSL performance by simply replacing the unseen expert with more powerful substitutes. 
\end{itemize}

%% file: sec/2_related.tex
\section{Related Work} 
\label{relatedwork}
\subsection{Audio-visual Generalized Zero-shot Learning}
Embedding approaches mapping the videos, audio signals, and text information into a shared common space to get joint feature representations aimed for subsequent classification or retrieval tasks enjoy wide popularity in AV-GZSL. Among them, prior CJME~\cite{cjme} employs triplet loss to restrict the distance between audio-visual features and class embedding on the proposed AudioSetZSL~\cite{cjme} dataset. Recently, TCaF~\cite{temporal} proposes a temporal cross-attention framework enhanced from AVCA~\cite{avca} by factoring in the temporal information. 
Hyperbolic~\cite{hyperbolic} method employs hyperbolic alignment loss and cross-attention module to further improve the separability of joint feature representations. EZ-AVGZL~\cite{ezavgzl} utilizes class embedding optimization to achieve better discriminability of class embeddings while maintaining their original semantics. ClipClap~\cite{clipclap} enhances the feature quality by pre-trained CLIP~\cite{clip} and CLAP~\cite{clap} models. As for the generative method, AVFS~\cite{avfs}(Audio-Visual Feature Synthesis) generates unseen samples to facilitate contrastive training.


\subsection{Out-of-distribution Detection and Post-hoc Methods}
Out-of-distribution~(OOD) detection is proposed to secure the smooth deployment of machine learning models in real-world scenarios since these models are typically trained and validated in close-world settings. OOD detection is implemented with different OOD scores~(scalars) produced by ID~(in-distribution) and OOD samples on which certain thresholds are applied to finish ID-OOD separation. 

 Post-hoc methods for OOD detection typically involve training a model or a classifier with ID data first and subsequently, the model with frozen parameters is converted into an OOD detector during the test stage, which is cost-effective compared with training an extra detector.
 Post-hoc methods compute an OOD score based on the pre-trained model's output. One simple solution MSP~\cite{msp} utilizes the maximum softmax probability as the OOD score to distinguish ID and OOD samples. To improve the discriminability of the softmax score, ODIN~\cite{odin} employs temperature scaling to make the softmax score distribution more uniform and also applies input perturbation. 
 Logits-based Energy score~\cite{energy, mood} makes use of output logits in conjunction with the \textbf{LogSumExp} function. Features-based methods~\cite{ash, knn, mos} like WDiscOOD~\cite{wdiscood} leverages LDA~(linear discriminant analysis)~\cite{lda} to enlarge the inter-class discrepancy and reduce the intra-class gap for better ID-OOD separation.

%% file: sec/3_method.tex
\section{Proposed Approach}
\label{method}
\begin{figure*}[!ht]
    \centering
    \includegraphics[width=7in]{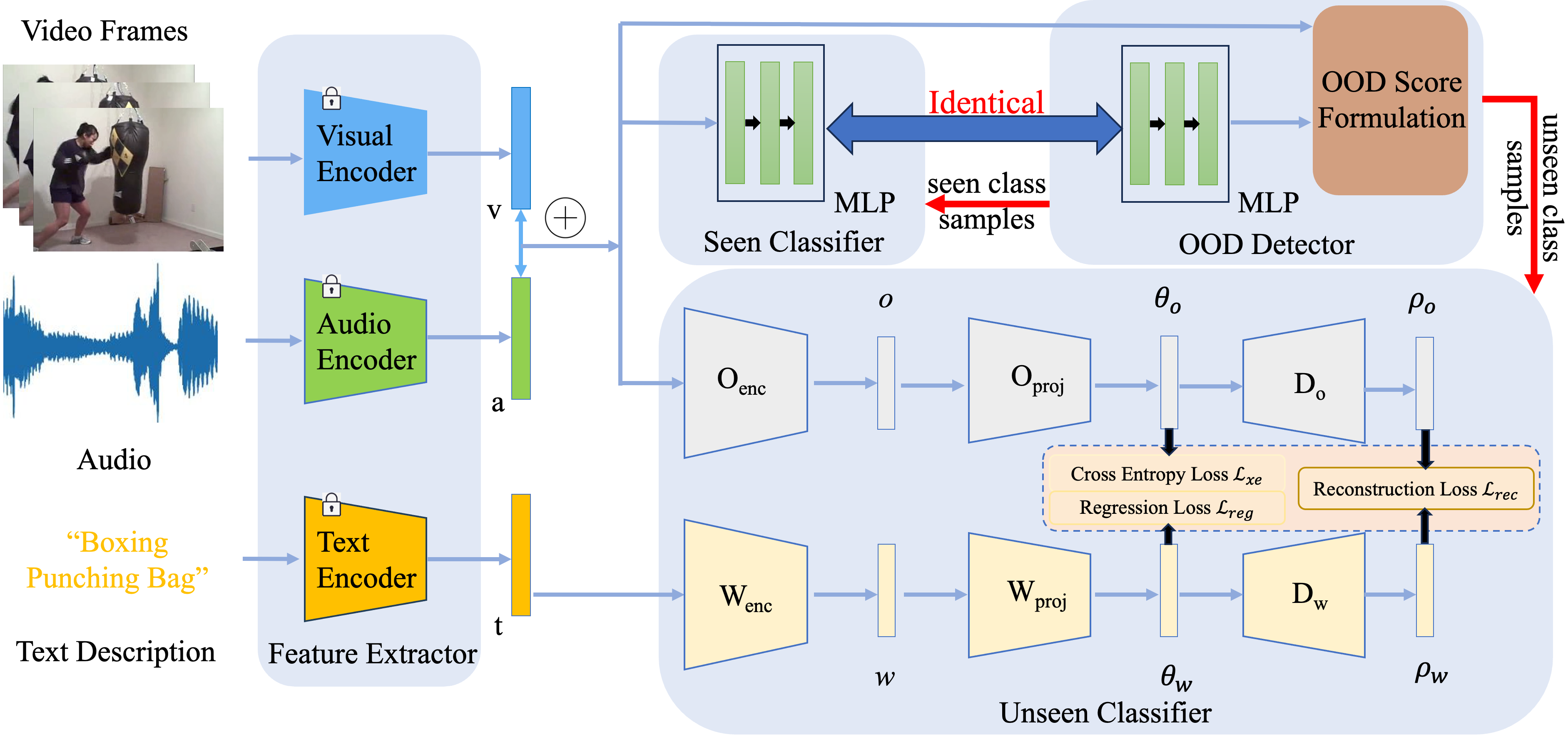}
    \caption{The general framework of EZ-AVOOD. Four key modules ``Feature Extractor'', ``OOD Detector'', ``Seen Classifier'' and ``Unseen Classifier'' make up the complete model. Parameter-fixed feature extractor simply produces audio-visual features $\boldsymbol{a} \oplus \boldsymbol{v}$~($\oplus$ represents concatenation operation) and text embeddings $\boldsymbol{t}$ without further optimization. Seen classifier and OOD detector are implemented with two identical MLPs, which means they share the same copy of parameters and need to train only one of them to make two modules work. The process of OOD score formulation is illustrated in Figure \ref{score}. 
    At evaluation stage, OOD detector distinguishes seen and unseen samples and input them to the trained seen expert and unseen expert classifiers respectively~(red arrows).}
    \label{framework}
\end{figure*}

\subsection{Problem Statement of Audio-visual GZSL}
AV-GZSL aims to efficiently recognize previously seen and even unseen video-audio combinations through the set of seen~(training) audio-visual events together with human-readable text descriptions. Thus, we denote samples from seen classes with 
$\boldsymbol{S}=(\boldsymbol{a}_i^s, \boldsymbol{v}_i^s, \boldsymbol{t}_i^s, y_i^s)_{i\in \{1,\cdots,M\}}$
, where $M$ seen samples in total consist of audio features $\boldsymbol{a}_i^s$, visual features $\boldsymbol{v}_i^s$ and textual description $\boldsymbol{t}_i^s$ as well as the corresponding ground truth class label $y_i^s$. 
Likewise, unseen dataset is denoted as 
$\boldsymbol{U}=(\boldsymbol{a}_j^u, \boldsymbol{v}_j^u, \boldsymbol{t}_j^u, y_j^u)_{j\in \{1,\cdots,K\}}$ 
with $K$ samples and notably 
${\boldsymbol{S}} \cap {\boldsymbol{U}} = \varnothing$. 
The number of seen classes labels is 
$C_s$: ${\boldsymbol{Y}^s} = { ({y_1^s}, \cdots ,{y}_M^s)} \in {\{{1, \cdots, C_s}\}}$
, and the number of unseen classes is 
$C_u$: ${\boldsymbol{Y}^u} = {({y_1^u}, \cdots ,{y}_K^u)} \in {\{{1, \cdots, C_u}\}}$. Hence,for the ZSL task, AV-GZSL learns the model $f_{ZSL}:\boldsymbol{X}\rightarrow{{{\boldsymbol{Y}}^{u}}}$ 
to classify unseen samples only, where 
$\boldsymbol{X} = (\boldsymbol{a}_z, \boldsymbol{v}_z)$ 
denotes the test dataset and in GZSL, the classifier $f_{GZSL}:\boldsymbol{X}\rightarrow{\boldsymbol{Y}^u \cup \boldsymbol{Y}^s}$ 
aims to classify seen and unseen examples.

\subsection{Model Architecture} \label{modelarch}
The proposed model is depicted in Figure \ref{framework} and there are three essential components in our model: the OOD detector, the supervised seen classifier, and the unseen classifier adapted from an existing embedding method. 
The greatest strength of our method is that there is no need to train a new OOD detector, since our OOD detector shares the same parameters with the seen classifier, which means the supervised classifier trained with seen samples also serves as the OOD detector. Therefore, EZ-AVOOD significantly reduces the complexity of the model and brings a considerable decrease in the computational overhead and training time. 

\subsubsection{Seen Classifier}
To cope with seen samples, a vanilla and efficient 3-layer MLP optimized with Cross Entropy Loss $\boldsymbol{\mathcal{L}}_{xent} = \textbf{CrossEntropy}(\boldsymbol{x}, y(\boldsymbol{x}))$ is adopted as the seen expert classifier, where $\boldsymbol{x}$ denotes the joint audio-visual features from seen classes constructed through simple concatenation: $\boldsymbol{x} = \boldsymbol{a} \oplus \boldsymbol{v}$. 
Moreover, once the seen classifier is trained, it can be leveraged as the OOD detector, and more details are thoroughly elaborated in the next part.

\begin{figure}[!ht]
    \centering
    \includegraphics[width=0.99\linewidth]{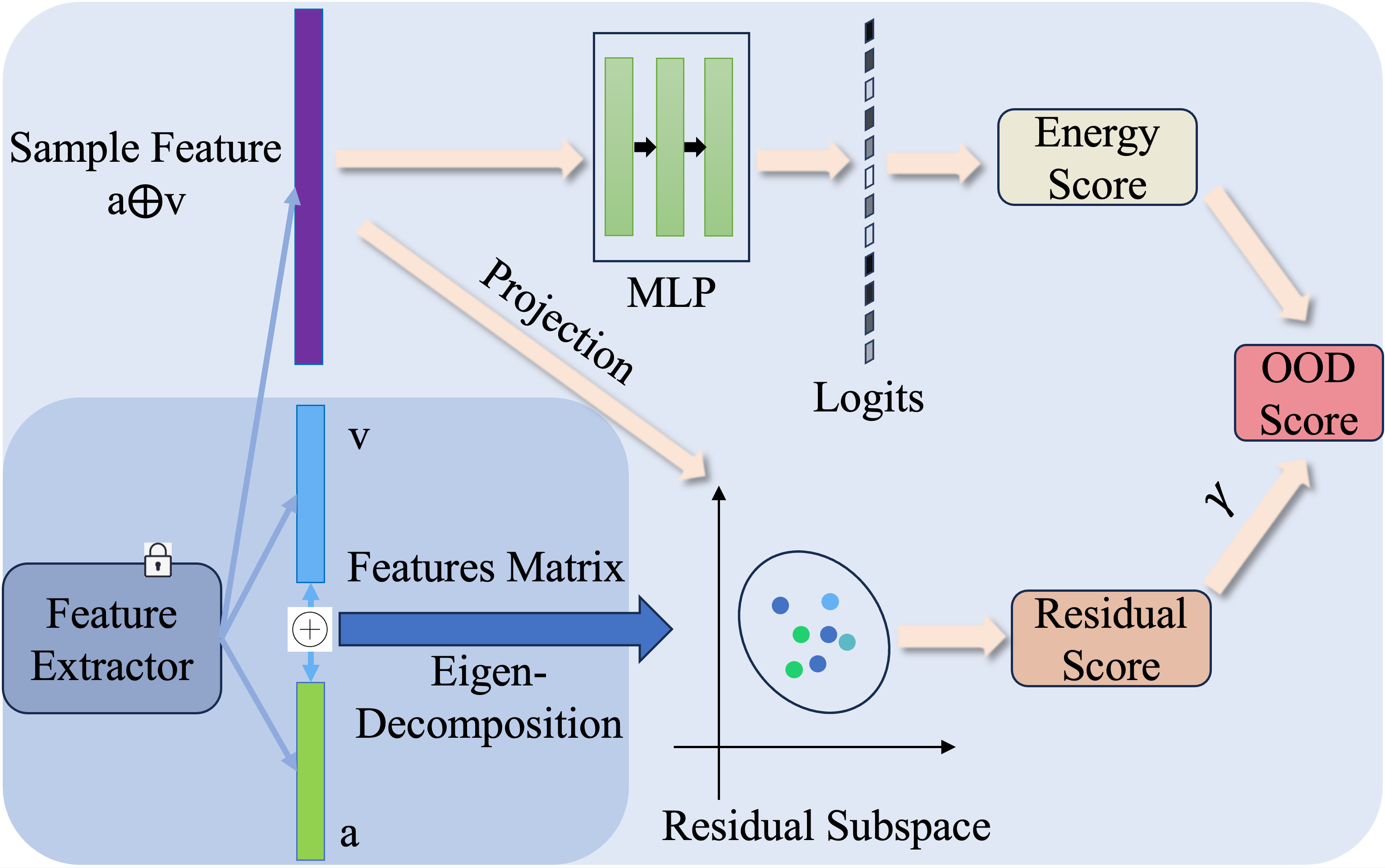}
    \caption{The process of the EZ-OOD score formulation. During training phase, ``Residual Subspace'' is derived from the eigen-decomposition on all seen samples features matrix. At test time~(pink arrows), concatenated audio-visual feature $\boldsymbol{a} \oplus \boldsymbol{v}$ projects onto the residual subspace to get ``Residual Score'' and ``Energy Score'' is calculated with the logits of the test sample produced by the MLP~(the trained seen classifier actually). The final OOD score is defined by the weighted sum of energy score and residual score.}
    \label{score}
\end{figure}

\subsubsection{Out-of-distribution Detector} \label{ood}
We adopt post-hoc idea to design OOD algorithm tackling seen-unseen separation in AV-GZSL problem. Since the output of pre-trained model usually includes high-dimensional features, logits, or Softmax probability and we choose to exploit the intrinsic information held in \textbf{class-specific logits} and \textbf{feature representation} to construct OOD score. Consequently, the trained seen classifier now becomes the ``OOD detector'' in our method to output class-dependent logits. The proposed extremely simple OOD detection method is named as ``EZ-OOD'' and the formulation pipeline is illustrated in Figure \ref{score}. The EZ-OOD score consists of Energy Score calculated by logits and Residual Score derived from residual subspace.

\textbf{Class-specific Logits and Energy Score} \quad
Seen classifier takes the high-dimensional fused audio-visual features as input and produces the logits corresponding to specific seen classes labels. We adopt the widely used Energy Score function $\textbf{LogSumExp}$ $\boldsymbol{E(x;\,l)}$ as one vital part of our OOD score, mapping the logits $\boldsymbol{l}$ of sample $\boldsymbol{x}$ to a scalar:
\begin{equation}
    \boldsymbol{E}(\boldsymbol{x};\,\boldsymbol{l}) = -\log\sum_{i=1}^C e^{l_i(\boldsymbol{x})}~,
\end{equation}
where $C$ is the number of seen classes and $l_i(\boldsymbol{x})$ is the logit of class-$i$ in correspondence with sample $\boldsymbol{x}$. Negative inversion $\boldsymbol{E}(\boldsymbol{x}) = -\boldsymbol{E}(\boldsymbol{x};\,\boldsymbol{l})$ is the score practically used to ensure that seen samples produce higher scores, which is consistent with the tradition in OOD detection. 

\textbf{Feature Representation and Class-agnostic Residual Score} \quad
Here $\boldsymbol{x} \in {\mathbb{R}}^D$ is the D-dimensional fused bi-modal sample feature and $\boldsymbol{X}$ denotes the audio-visual feature matrix of all seen samples. Therefore, the principal subspace ${P}$ is defined by the $N$-dimensional space spanned by the eigenvetors corresponding to the top-$N$ eigenvalues of matrix $\boldsymbol{X}^T\boldsymbol{X}$ and the the residual subspace ${P}^{\perp}$ is the orthogonal complements of the space ${P}$. Thus we have $\boldsymbol{x} = \boldsymbol{x}^P + \boldsymbol{x}^{{P}^{\perp}}$ where $\boldsymbol{x}^P$ is the projection of sample feature $\boldsymbol{x}$ onto subspace $P$ and $\boldsymbol{x}^{{P}^{\perp}}$ is the mapping on ${P}^{\perp}$. Suppose the eigen-decomposition of matrix $\boldsymbol{X}^T\boldsymbol{X}$ is
\begin{equation}
    \boldsymbol{X}^T\boldsymbol{X} = \boldsymbol{W{\Lambda}W}^T~,
\end{equation}
where $\boldsymbol{W}$ refers to a set of standard orthogonal bases that are arranged according to the decreasing order of the eigenvalues within the diagonal matrix $\boldsymbol{\Lambda}$. $N$-dimensional principal subspace $P$ is defined by the first $N$ column vectors of $\boldsymbol{W}$, and span of the $(N+1)$-th column to the $D$-th column vectors in $\boldsymbol{W}$ is the residual subspace ${P}^{\perp}$. Then the matrix $\boldsymbol{W}$ can be separated into matrix $\boldsymbol{Q} \in {\mathbb{R}}^{D\times{N}}$ and matrix $\boldsymbol{O} \in {\mathbb{R}}^{D\times(D-N)}$ formed by the last $(D-N)$ eigenvetors, and we can get
\begin{equation}
    \boldsymbol{x}^P = {\boldsymbol{Q}}^T\boldsymbol{x}; \ \boldsymbol{x}^{{P}^{\perp}} = {\boldsymbol{O}}^T\boldsymbol{x}~.
\end{equation}

 Given that the principal subspace and the residual subspace are constructed based on the feature representations of all training samples, they just ignore the information specific to individual seen category, namely characteristics hidden in these two subspaces are \textbf{class-agnostic}. Moreover, we argue that seen samples are relatively closer to principal subspace while deviate a lot from residual subspace. Therefore we define the Residual Score $bm{R}(\boldsymbol{x})$ as the norm of $\boldsymbol{x}^{{P}^{\perp}}$:
 \begin{equation}
    \boldsymbol{R}(\boldsymbol{x}) = - \|\boldsymbol{x}^{{P}^{\perp}} \| = -\left(\boldsymbol{x}^T{\boldsymbol{O}}{\boldsymbol{O}}^T\boldsymbol{x}\right)^{1/2}~.
\end{equation}

Just like the Energy Score, we take a minus norm to make ID samples produce higher OOD scores.  

\textbf{EZ-OOD Score Formulation} \quad
 The final OOD score $\boldsymbol{S}(\boldsymbol{x})$ is formulated by the weighted sum of energy score and residual score to unify the \textbf{class-specific} and \textbf{class-agnostic} information for better ID-OOD separation:
\begin{equation}
    \boldsymbol{S}(\boldsymbol{x}) = \boldsymbol{E}(\boldsymbol{x}) + \gamma \boldsymbol{R}(\boldsymbol{x})~,
\end{equation}
where $\gamma$ is the weight hyper-parameter to balance the scale of these two different scores and enhance the overall OOD detection performance.

OOD detection process is defined as below, $A$ is the binary classification outcome,
\begin{equation}
A_\lambda(\boldsymbol{x})= 
\begin{cases}
\text { Seen } & \boldsymbol{S}(\boldsymbol{x}) \geq \lambda \\ 
\text { Unseen } & \boldsymbol{S}(\boldsymbol{x})<\lambda ~,
\end{cases}
\end{equation}
where $\lambda$ is the threshold and samples possess higher $\boldsymbol{S}(\boldsymbol{x})$ are tend to be treated as seen classes. The threshold is uniquely determined by the training samples and has nothing to do with the test data.

\begin{table*}[!ht]
\centering
\resizebox{\textwidth}{!}{
\begin{tabular}{ccccccccccccc}
\toprule
\multirow{2}{*}{Methods} & \multicolumn{4}{c}{VGGSound-GZSL$^{cls}$} & \multicolumn{4}{c}{UCF-GZSL$^{cls}$} & \multicolumn{4}{c}{ActivityNet-GZSL$^{cls}$} \\
\cmidrule(r){2-5} \cmidrule(r){6-9} \cmidrule(r){10-13}
& $acc_{\boldsymbol{S}}$ & $acc_{\boldsymbol{U}}$ & $\boldsymbol{H}$ & $acc_{ZSL}$ & $acc_{\boldsymbol{S}}$ & $acc_{\boldsymbol{U}}$ & $\boldsymbol{H}$ & $acc_{ZSL}$ & $acc_{\boldsymbol{S}}$ & $acc_{\boldsymbol{U}}$ & $\boldsymbol{H}$ & $acc_{ZSL}$\\ 
\midrule
CJME~\cite{cjme} & 11.96 & 5.41 & 7.45 & 6.84 & 48.18 & 17.68 & 25.87 & 20.46 & 16.06 & 9.13 & 11.64 & 9.92\\
AVGZSLNet~\cite{AVGZSLNet} & 13.02 & 2.88 & 4.71 & 5.44 & 56.26 & 34.37 & 42.67 & 35.66 & 14.81 & 11.11 & 12.70 & 12.39\\
AVCA~\cite{avca} & \underline{32.47} & 6.81 & 11.26 & 8.16 & 34.90 & 38.67 & 36.69 & 38.67 & 24.04 & 19.88 & 21.76 & 20.88\\
Hyper-multiple~\cite{hyperbolic} & 21.99 & 8.12 & 11.87 & 8.47 & 43.52 & 39.77 & 41.56 & 40.28 & 20.52 & \textbf{21.30} & 20.90 & 22.18\\
ClipClap~\cite{clipclap} & 29.68 & \underline{11.12} & \underline{16.18} & \underline{11.53} & \underline{77.14} & \underline{43.91} & \underline{55.97} & \underline{46.96} & \textbf{45.98} & 20.06 & \underline{27.93} & \underline{22.76}\\
\midrule
\textbf{EZ-AVOOD}~(\textbf{Ours}) & \textbf{39.33} & \textbf{11.84} & \textbf{18.21} & \textbf{13.28} & \textbf{83.53} & \textbf{48.01} & \textbf{60.97} & \textbf{50.92} & \underline{41.56} & \underline{21.06} & \textbf{27.95} & \textbf{25.20}\\
\bottomrule  
\end{tabular}}
\caption{Comparison with existing state-of-the-art methods on VGGSound-GZSL$^{cls}$, UCF-GZSL$^{cls}$ and ActivityNet-GZSL$^{cls}$ datasets. Performances in percentage of GZSL~($acc_{\boldsymbol{S}}$/$acc_{\boldsymbol{U}}$/H) and ZSL~($acc_{ZSL}$)  are reported. For fair comparison, results of all five baseline methods are obtained using audio-visual features and text embeddings extracted by CLIP and CLAP models. Bold values represent the best results and the second-ranked numbers are underlined.}
\label{1}
\end{table*}

\subsubsection{Unseen Classifier}
Here we fine-tune the ClipClap~\cite{clipclap} model to enhance the unseen classes average accuracy for the purpose of improving final GZSL performance. As illustrated in Figure \ref{framework}, the general framework consists of two branches of Encoder-Encoder-Decoder pipeline to deal with concatenated audio-visual features and fused text embeddings, respectively. With respect to the feature extraction, to be specific, visual features $\boldsymbol{v}$ and part of concatenated text embeddings $\boldsymbol{t^v}$ are extracted by vision-language pre-trained model CLIP~\cite{clip} and CLAP~\cite{clap} model dedicated to audio-language tasks produces audio features $\boldsymbol{a}$ and another part of text embeddings $\boldsymbol{t^a}$. 

The first encoder block $\boldsymbol{O}_{enc}$ from audio-visual branch takes concatenated features $\boldsymbol{x} = \boldsymbol{a} \oplus \boldsymbol{v}$ as input and outputs the multimodal sample features $\boldsymbol{o}$ :
\begin{equation}
    \boldsymbol{o} = \boldsymbol{O}_{enc}(\boldsymbol{x})~.
\end{equation}

In the same way, we get unified text embeddings $\boldsymbol{w}$ with encoder $\boldsymbol{W}_{enc}$:
\begin{equation}
    \boldsymbol{w} = \boldsymbol{W}_{enc}(\boldsymbol{t^a} \oplus \boldsymbol{t^v} )~.
\end{equation}

With multimodal sample features $\boldsymbol{o}$ and fused text embeddings $\boldsymbol{w}$ as the input of the rest two simple and effective Encoder-Decoder compound modules, we have the following formulations:
\begin{equation}
    \boldsymbol{\theta}_o = \boldsymbol{O}_{proj}(\boldsymbol{o}); \ \boldsymbol{\rho}_o = \boldsymbol{D}_o(\boldsymbol{\theta}_o)~,
\end{equation}
\begin{equation}
    \boldsymbol{\theta}_w = \boldsymbol{W}_{proj}(\boldsymbol{w}); \ \boldsymbol{\rho}_w = \boldsymbol{D}_w(\boldsymbol{\theta}_w)~,
\end{equation}
where $\boldsymbol{\theta}_o$ and $\boldsymbol{\theta}_w$ represent the projection outcomes, while the reconstruction process produces $\boldsymbol{\rho}_o$ and $\boldsymbol{\rho}_w$. Then the training objectives of unseen classifier include Cross Entropy Loss $\boldsymbol{\mathcal{L}}_{xe}$:
\begin{equation}
\boldsymbol{\mathcal{L}}_{xe}=-\frac{1}{n} \sum_i^n \log \left(\frac{\exp \left(\boldsymbol{\theta}_{w_{y_{i}^s}} \boldsymbol{\theta}_{o_i}\right)}{\sum_{j}^{C_s} \exp \left(\boldsymbol{\theta}_{w_{y_j^s}} \boldsymbol{\theta}_{o_i}\right)}\right)~,
\end{equation}
where $y_{i}^s$ is the label of seen sample $i$, and $\boldsymbol{\theta}_{w_{y_{i}^s}}$ denotes the $\boldsymbol{\theta}_w$-projection of text embedding belonging to seen class $y_i^s$. $n$ and $C_s$ represent the number of training samples and seen categories, respectively. Another loss function is Reconstruction Loss $\boldsymbol{\mathcal{L}}_{rec}$ to minimize the discrepancy between $\boldsymbol{\rho}$ and text embeddings $\boldsymbol{w}$ with MSE~(mean squared error):
\begin{equation}
    \boldsymbol{\mathcal{L}}_{rec} = \frac{1}{n} \sum_i^n \left[ {\left(\boldsymbol{\rho}_{o_i} - \boldsymbol{w}_i\right)}^2 +  {\left(\boldsymbol{\rho}_{w_i} - \boldsymbol{w}_i\right)}^2\right]~.
\end{equation}

Moreover, a Regression Loss $\boldsymbol{\mathcal{L}}_{reg}$ calculated by MSE function between $\boldsymbol{\theta}_o$ and $\boldsymbol{\theta}_w$ is defined as:
\begin{equation}
    \boldsymbol{\mathcal{L}}_{reg} = \frac{1}{n} \sum_i^n {\left(\boldsymbol{\theta}_{o_i} - \boldsymbol{\theta}_{w_i}\right)}^2~.
\end{equation}

And the overall loss for unseen classifier is defined as:
\begin{equation}
    \boldsymbol{\mathcal{L}}_{total} = \boldsymbol{\mathcal{L}}_{xe} + \boldsymbol{\mathcal{L}}_{rec} + \boldsymbol{\mathcal{L}}_{reg}~.
\end{equation}

Following the original loss function design, there is no weight parameter applied on final loss $\boldsymbol{\mathcal{L}}_{total}$. 

During test phase, the classification result is determined by the nearest neighbor principle which means class text embedding closest to sample feature projection $\boldsymbol{\theta}_o^i$ is selected as the predicted label $c_i$:
\begin{equation}
c_i=\underset{j}{\operatorname{argmin}}\left(\left\|\boldsymbol{\theta}_w^j-\boldsymbol{\theta}_o^i\right\|_2\right)~,
\end{equation}
where $\boldsymbol{\theta}_w^j$ is the encoded text embeddings corresponding to class $i$.

\begin{table*}[!ht]
\centering
\resizebox{\textwidth}{!}{
\begin{tabular}{ccccccccccccc}
\toprule
\multirow{2}{*}{Methods} & \multicolumn{4}{c}{VGGSound-GZSL$^{main}$} & \multicolumn{4}{c}{UCF-GZSL$^{main}$} & \multicolumn{4}{c}{ActivityNet-GZSL$^{main}$} \\
\cmidrule(r){2-5} \cmidrule(r){6-9} \cmidrule(r){10-13}
& $acc_{\boldsymbol{S}}$ & $acc_{\boldsymbol{U}}$ & $\boldsymbol{H}$ & $acc_{ZSL}$ & $acc_{\boldsymbol{S}}$ & $acc_{\boldsymbol{U}}$ & $\boldsymbol{H}$ & $acc_{ZSL}$ & $acc_{\boldsymbol{S}}$ & $acc_{\boldsymbol{U}}$ & $\boldsymbol{H}$ & $acc_{ZSL}$\\ 
\midrule
CJME~\cite{cjme} & 8.69 & 4.78 & 6.17 & 5.16 & 26.04 & 8.21 & 12.48 & 8.29 & 5.55 & 4.75 & 5.12 & 5.84 \\
AVGZSLNet~\cite{AVGZSLNet} & 18.15 & 3.48 & 5.83 & 5.28 & 52.52 & 10.90 & 18.05 & 13.65 & 8.93 & 5.04 & 6.44 & 5.40 \\
TCaF~\cite{temporal} & 9.64 & 5.91 & 7.33 & 6.06 & 58.60 & 21.74 & 31.72 & 24.81 & 18.70 & 7.50 & 10.71 & 7.91 \\
VIB-GZSL~\cite{vib} & \underline{18.42} & 6.00 & 9.05 & 6.41 & \textbf{90.35} & 21.41 & 34.62 & 22.49 & 22.12 & 8.94 & 12.73 & 9.29 \\
AVMST~\cite{avmst} & 14.14 & 5.28 & 7.68 & 6.61 & 44.08 & 22.63 & 29.91 & 28.19 & 17.75 & \underline{9.90} & 12.71 & 10.37 \\
MDFT~\cite{mdft} & 16.14 & 5.97 & 8.72 & 7.13 & 48.79 & 23.11 & 31.36 & \textbf{31.53} & 18.32 & \textbf{10.55} & 13.39 & \textbf{12.55} \\
Hyper-multiple~\cite{hyperbolic} & 15.02 & \underline{6.75} & \underline{9.32} & \textbf{7.97} & 63.08 & 19.10 & 29.32 & 22.24 & 23.38 & 8.67 & 12.65 & 9.50 \\
\midrule 
AVCA~\cite{avca} & 14.90 & 4.00 & 6.31 & 6.00 & 51.53 & 18.43 & 27.15 & 20.01 & 24.86 & 8.02 & 12.13 & 9.13 \\
OOD-entropy+AVCA~\cite{avood} & 13.31 & \textbf{7.01} & 9.19 & \underline{7.48} & 63.94 & \underline{26.99} & \underline{37.96} & \underline{30.56} & \underline{29.84} & 9.54 & \textbf{14.46} & \underline{11.41}\\
\textbf{EZ-OOD}+\textbf{AVCA}~(\textbf{Ours}) & \textbf{24.94} & 6.38 & \textbf{10.16} & \underline{7.48} & \underline{79.71} & \textbf{27.94} & \textbf{41.38} & \underline{30.56} & \textbf{30.65} & 9.29 & \underline{14.26} & \underline{11.41} \\
\bottomrule  
\end{tabular}}
\caption{Compatibility of EZ-OOD with existing method. We make comparisons between existing state-of-the-art AV-GZSL methods and our new model on VGGSound-GZSL$^{main}$, UCF-GZSL$^{main}$, and ActivityNet-GZSL$^{main}$ datasets. GZSL~($acc_{\boldsymbol{S}}$/$acc_{\boldsymbol{U}}$/$\boldsymbol{H}$) and ZSL~($acc_{ZSL}$) performances are reported in percentage. Bold numbers denote the best results and the second highest values are underlined. 
}
\label{2}
\end{table*}

%% file: sec/4_experiment.tex
\section{Experiments and Results Analysis} 
\label{exp}
\subsection{Setup for Audio-visual GZSL}
\subsubsection{Datasets and Evaluation Metrics}
Following experimental setting in AVCA~\cite{avca}, we adopt the curated version of three audio-visual datasets: 
VGGSound~\cite{vggsound}, UCF101~\cite{ucf101}, and ActivityNet~\cite{activitynet} to evaluate our EZ-AVOOD method and they are VGGSound-GZSL$^{cls}$, UCF-GZSL$^{cls}$ and ActivityNet-GZSL$^{cls}$. The upper $cls$ represents $cls$-$split$ introduced in \cite{temporal} instead of $main$-$split$ utilized in \cite{avca}. 

Consistent with ZSL conventions~\cite{xian2018zero,avca}, we adopt the average per-class classification accuracy as the evaluation metric, where $acc_{\boldsymbol{S}}$ and $acc_{\boldsymbol{U}}$ denotes the mean class accuracy of seen classes and unseen classes separately. To comprehensively evaluate GZSL performance, the harmonic mean $\boldsymbol{H}$ of seen and unseen accuracy is calculated as: 
\begin{equation}
    \boldsymbol{H} = \frac{2 * acc_{\boldsymbol{S}} * acc_{\boldsymbol{U}}}{acc_{\boldsymbol{S}} + acc_{\boldsymbol{U}}}~.
\end{equation}

For ZSL tasks aimed to classify unseen samples only, mean class accuracy $acc_{ZSL}$ is also obtained. 

\subsubsection{Implementation Details}
\textbf{OOD Detector} \quad The trained seen classifier is transformed into our OOD detector to produce logits and energy score. As for residual score, the dimension N of principal subspace and the scaling factor $\gamma$ are valued at 64/90 for VGGSound-GZSL$^{cls}$, 256/205 for UCF-GZSL$^{cls}$ and 256/285 for ActivityNet-GZSL$^{cls}$.

More details about Feature Extractor, Seen Classifier, and Unseen Classifier are provided in supplementary material.

\subsection{Experimental Results}\label{experiment}
\subsubsection{Quantitative Results}
As shown in Table \ref{1}, our EZ-AVOOD model consistently takes the lead on all three benchmarks in terms of both harmonic mean $\boldsymbol{H}$ for GZSL task and $acc_{ZSL}$ under ZSL setup. For VGGSound-GZSL$^{cls}$ dataset, we achieve the best performances on all metrics and specially EZ-AVOOD considerably outperforms current state-of-the-art ClipClap with the lead of 9.65\%@$acc_{\boldsymbol{S}}$, 0.72\%@$acc_{\boldsymbol{U}}$, 2.03\%@$\boldsymbol{H}$, and 1.75\%@$acc_{ZSL}$. In addition, our method substantially overtakes the ClipClap on UCF-GZSL$^{cls}$ benchmark with even bigger lead margins of 6.39\%@$acc_{\boldsymbol{S}}$, 4.10\%@$acc_{\boldsymbol{U}}$, 5.00\%@$\boldsymbol{H}$, and 3.96\%@$acc_{ZSL}$ respectively. Though the proposed EZ-AVOOD ``merely'' takes the second place on $acc_{\boldsymbol{S}}$ and $acc_{\boldsymbol{U}}$ of ActivityNet-GZSL$^{cls}$, in which ClipClap and Hyper-multiple ranks the top separately, our method does holds the first place on the more comprehensive metric $\boldsymbol{H}$ and attains significant performance on $acc_{ZSL}$ better than all other baseline methods with a 2.44\% lead margin at least.

\subsection{Compatibility of EZ-OOD with Existing Method}\label{compa}
\subsubsection{Experimental Setup}
Here, we replace the unseen classifier of EZ-AVOOD with AVCA~\cite{avca} and explore the new model's experimental performance. We provide some key details of this experiment: $main$-$split$ of three datasets is adopted; audio features and visual features are extracted by self-supervised SeLaVi~\cite{selavi} pre-trained on VGGSound dataset; and text embeddings are obtained using word2vec model~\cite{w2v} pre-trained with Wikipedia. 
\subsubsection{Baseline Method}
AV-OOD\cite{avood} is another OOD-based method that takes the AVCA model as unseen expert and proposes the OOD-entropy method for OOD detection, consequently quite suitable for making contrasts with our method right here. To ensure a just comparison, we only train the seen classifier~(also working as the OOD detector) and utilize the same unseen classifier
as AV-OOD method. Notably, we re-run the AV-OOD with the provided checkpoint files and get ZSL and GZSL performances to facilitate fair comparison. As can be seen in Table \ref{2}, the $acc_{ZSL}$ results of our method and OOD-entropy are totally the same on three benchmarks. 

\begin{table*}[!ht]
\centering
\resizebox{\textwidth}{!}{
\begin{tabular}{ccccccccccccc}
\toprule
\multirow{3}{*}{Methods} & \multicolumn{4}{c}{VGGSound-GZSL$^{cls}$} & \multicolumn{4}{c}{UCF-GZSL$^{cls}$} & \multicolumn{4}{c}{ActivityNet-GZSL$^{cls}$} \\
\cmidrule(r){2-5} \cmidrule(r){6-9} \cmidrule(r){10-13}
& AUROC & FPR95 & AUPR & $\boldsymbol{H}$ & AUROC & FPR95 & AUPR & $\boldsymbol{H}$  & AUROC & FPR95 & AUPR & $\boldsymbol{H}$  \\
& $\uparrow$ & $\downarrow$ & $\uparrow$  & $\uparrow$  & $\uparrow$ & $\downarrow$ & $\uparrow$  & $\uparrow$  & $\uparrow$ & $\downarrow$ & $\uparrow$  & $\uparrow$ \\
\midrule
Residual Score & 68.73 & 85.74 & 85.51 & 14.91 & \underline{91.30} & \underline{42.09} & \underline{90.96} & \underline{58.65} & 67.30 & 85.28 & 44.73 & 25.34 \\
Energy Score & \underline{81.97} & \underline{67.69} & \underline{92.65} & \underline{17.84} & 80.41 & 74.09 & 86.55 & 47.57 & \underline{70.99} & \underline{86.95} & \underline{54.57} & \underline{26.13}\\
\textbf{EZ-OOD~(full)} & \textbf{84.33} & \textbf{66.24} & \textbf{93.82} & \textbf{18.21} & \textbf{95.35} & \textbf{33.87} & \textbf{96.01} & \textbf{60.97} & \textbf{77.57} & \textbf{80.61} & \textbf{63.62} & \textbf{27.95} \\
\bottomrule  
\end{tabular}}
\caption{Ablation studies on EZ-OOD method. We make a comparison with Energy Score, Resdual Score, and their $\gamma$-weighted sum the full EZ-OOD on VGGSound-GZSL$^{cls}$, UCF-GZSL$^{cls}$ and ActivityNet-GZSL$^{cls}$ datasets. Out-of-distribution detection metrics~(AUROC/FPR95/AUPR) and GZSL~(harmonic mean $\boldsymbol{H}$) performance are reported in percentage. $\downarrow$ indicates that lower results are better while $\uparrow$ means the opposite. Bold values denote the best results and the second-best outcomes are underlined.}
\label{4}
\end{table*}

\begin{figure*}[!ht]
    \centering
    \includegraphics[width=6.8in]{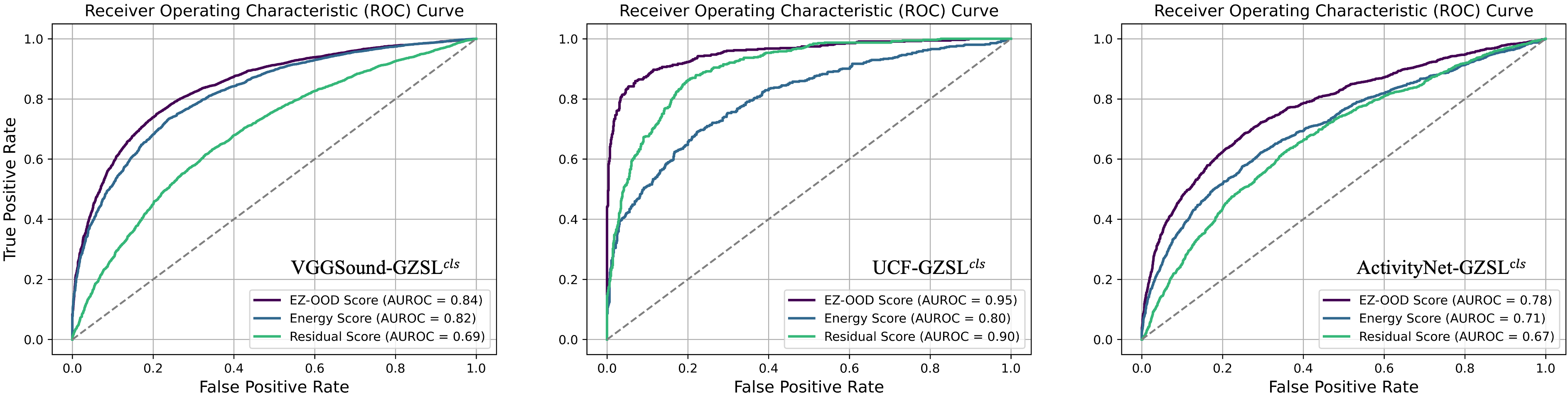}
    \caption{ROC curves of EZ-OOD, Energy Score, and Residual Score on three datasets. 
    Evidently, the full EZ-OOD consistently outperforms Energy Score and Residual Score with larger AUROC metric.}
    \label{ablation}
\end{figure*}

\subsubsection{Quantitative Results}
In the first place, compared with baseline method AVCA, the new model has secured a full-fledged lead on all metrics of three different datasets. More accurately, the greatest lead margins can amount to 28.18\%@$acc_{\boldsymbol{S}}$, 9.51\%@$acc_{\boldsymbol{U}}$, 14.23\%@$\boldsymbol{H}$, and 10.55\%@$acc_{ZSL}$ on UCF-GZSL$^{main}$ dataset. 
In addition, our new model significantly increases the ZSL and GZSL performances on the other 2 benchmarks from the baseline which comprehensively verifies the compatibility of the proposed EZ-OOD method.

Secondly, compared with OOD-entropy+AVCA~\cite{avood}, our EZ-OOD leveraging the same unseen classifier attains a remarkable lead on the harmonic mean metric of both VGGSound-GZSL$^{main}$ and UCF-GZSL$^{main}$ datasets~(0.97\% and 3.42\% separately), and lags behind by merely 0.2\%@$\boldsymbol{H}$ on ActivityNet-GZSL$^{main}$ benchmark. Notably, the OOD detection performance of our method is ahead of OOD-entropy actually on ActivityNet benchmark which is illustrated in the supplementary material. Since AV-GZSL evaluates the average per-class classification accuracy, while OOD detection simply considers ID-OOD separation of all test samples without caring about class labels, as a result, better but close OOD performance not always brings stronger GZSL performance.

\subsection{Ablation Studies}
To gain an insight into the concrete effect of Energy Score and Residual Score, we conduct additional ablation studies to compare the OOD detection performance and AV-GZSL results within EZ-OOD and its two key components. Experimental setup is consistent with the proposed EZ-AVOOD model in Experiment \ref{experiment}.
Here in Table \ref{4} we report the AUROC~(Area Under the Receiver Operating Characteristic curve), FPR95~(FPR@TPR95), and AUPR~(Area Under the Precision versus Recall curve) to evaluate OOD detection capability as well as the harmonic mean $\boldsymbol{H}$ of GZSL task on three datasets.
Also, we draw the ROC curves belonging to the three methods to explicitly display their OOD detection performance on each benchmark in Figure \ref{ablation}.

\begin{figure}[!ht]
    \centering
    \includegraphics[width=0.98\linewidth]{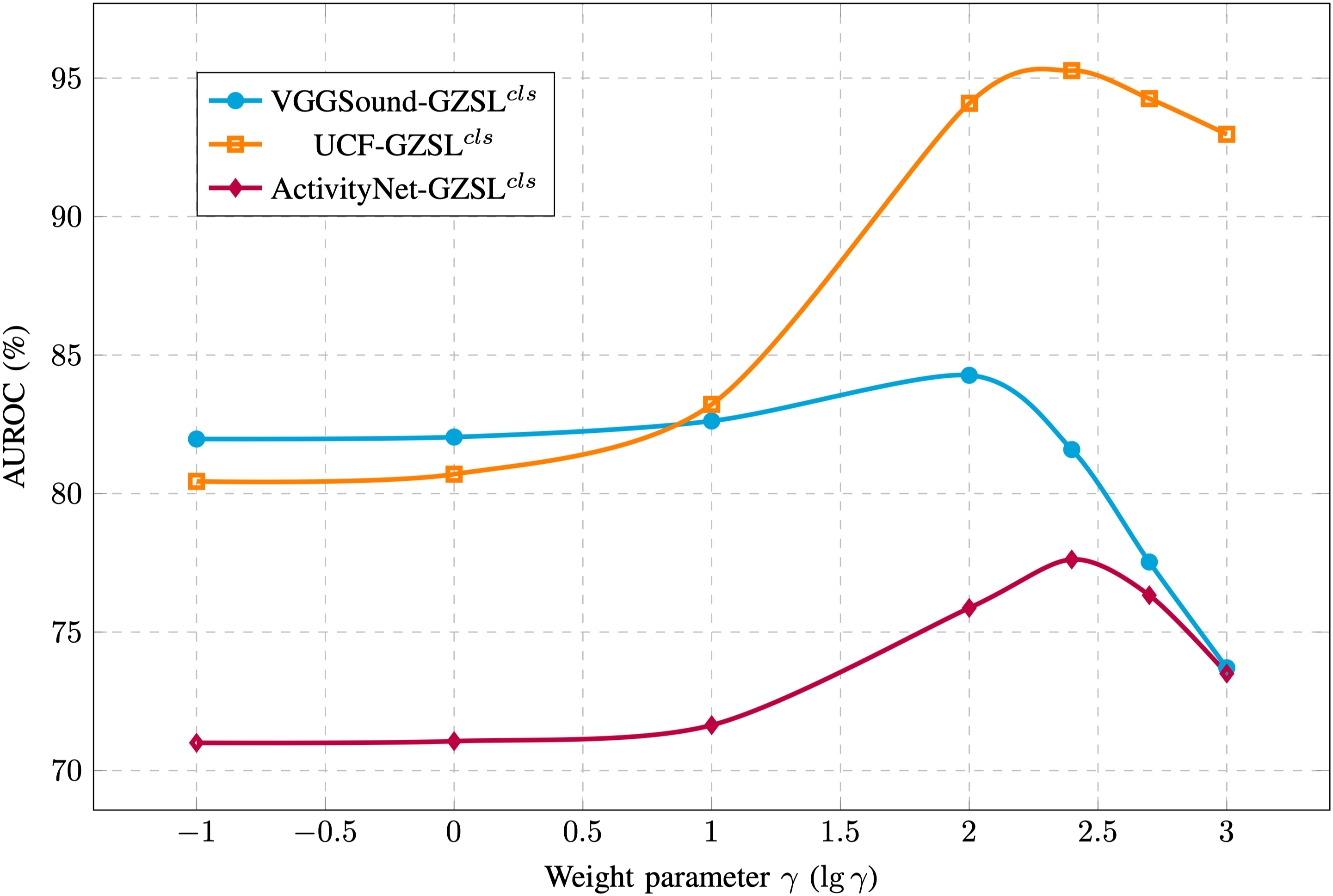}
    \caption{Effect of scaling factor $\gamma$ on AUROC for three datasets. OOD detection performance of EZ-OOD reaches the top when energy score and residual score are properly matched with the linear combination scaled by a suitable $\gamma$.}
    \label{gamma}
\end{figure}

\subsubsection{Quantitative Results and Qualitative Results}
According to the results in Table \ref{4}, the full EZ-OOD undoubtedly takes the first place on all three OOD detection metrics with the highest AUROC and AUPR and the lowest FPR95 and naturally achieves the best $\boldsymbol{H}$ for GZSL. In terms of the two components, Energy Score ranks second on VGGSound-GZSL$^{cls}$ and ActivityNet-GZSL$^{cls}$ datasets and Residual Score attains a remarkable lead over Energy Score on UCF-GZSL$^{cls}$ benchmark. Moreover, we observe that the harmonic mean $\boldsymbol{H}$ produce by Energy Score@VGGSound-GZSL$^{cls}$~(17.84\%) and Residual Score@UCF-GZSL$^{cls}$~(58.65\%) effortlessly defeat all the contrasting methods in Table \ref{1}. Therefore, we conclude that both Energy Score and Residual Score play a vital role in separating seen and unseen samples to facilitate subsequent ZSL classification objectives. 

Figure \ref{ablation} depicts the AUROC discrepancy between EZ-OOD score and its two crucial components on three benchmarks. In addition, two individual OOD scores manifest competitive detection performance whose ROC curves are close to the upper-left in the graph. To sum up, the proposed OOD score effectively combines the strengths of the two powerful scores with the weighted sum to achieve stronger OOD detection performance and higher GZSL classification accuracy.

\subsection{Parameter Sensitivity Studies}
\subsubsection{Effect of Scaling Factor \texorpdfstring{$\gamma$}{}}
Here we test the scaling factor $\gamma$ from the set: $\gamma \in \{ 0.1, 1, 10, 100, 250, 500, 1000\}$ with a fixed N specific to each dataset. Typically, better OOD detection performance will bring higher $\boldsymbol{H}$ for GZSL task, hence, AUROC is evaluated to avoid extra computational burden instead of $\boldsymbol{H}$. We follow the same experimental setup in Section \ref{experiment}. Figure \ref{gamma} illustrates the AUROC on three datasets with different scaling factors $\gamma$. When the scaling factor is valued at 0.1 or 1000, the linear combination EZ-OOD score will reduce to ordinary energy score or individual residual score, resulting in lower AUROC at both ends of the curve, which is consistent with the outcomes in ablation studies. Moreover, our method is capable of effectively integrating the discriminative information held by class-specific energy score and class-agnostic residual score under a wide range of scaling factors to achieve enhanced OOD detection performance and better audio-visual GZSL results than individual scores. 

\begin{figure}[!ht]
    \centering
    \includegraphics[width=0.98\linewidth]{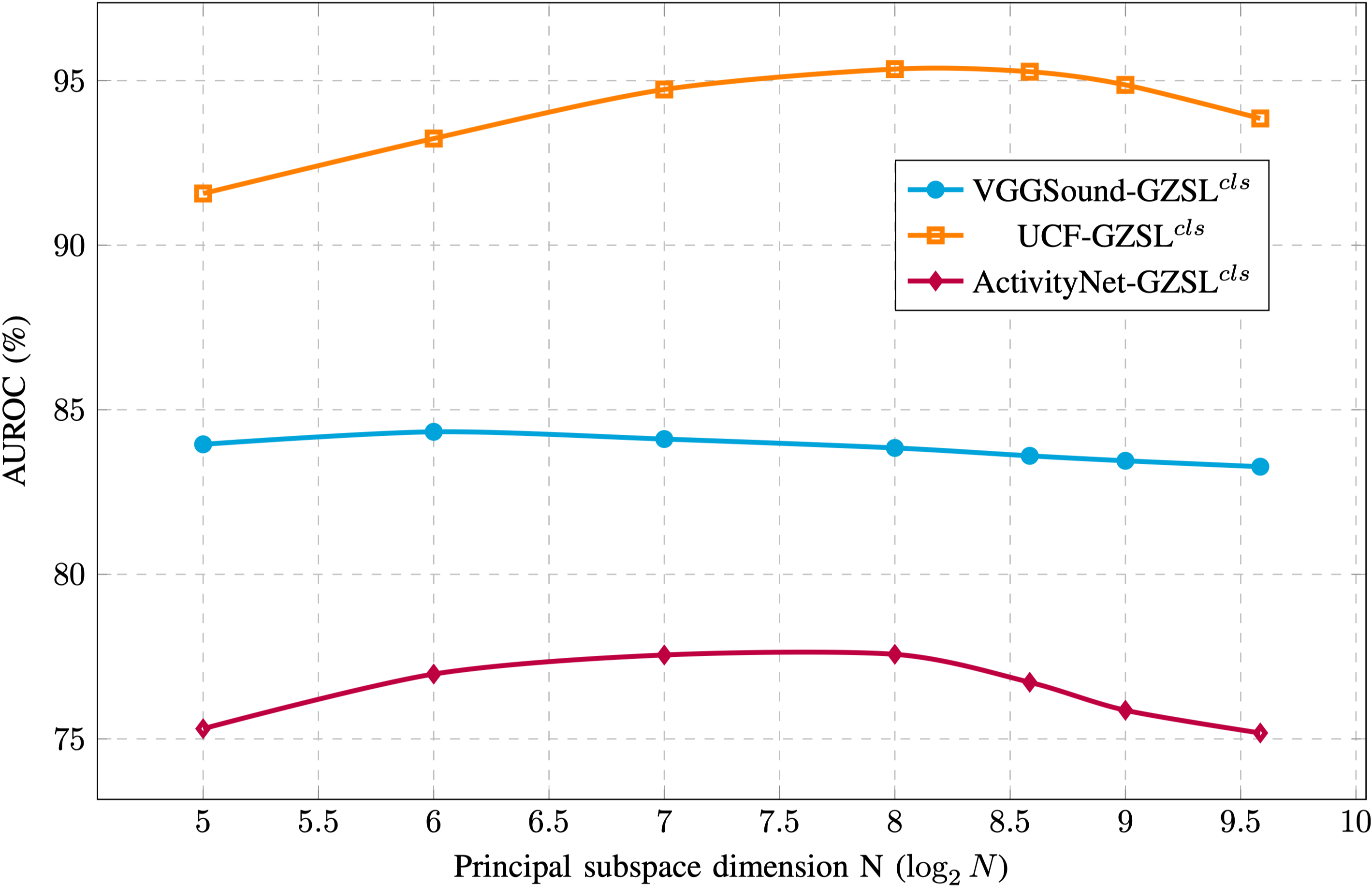}
    \caption{Effect of principal subspace dimension N on AUROC for three datasets. OOD detection performance of EZ-OOD method is quite robust with principal subspace dimension since the AUROC changes little towards a wide range of N values.}
    \label{dimension}
\end{figure}

\subsubsection{Effect of Principle Subspace Dimension N}
Different N parameters will have a direct influence on the OOD detection performance of residual score, followed by producing different EZ-OOD scores and finally change the overall audio-visual GZSL results. Additionally, since the concatenated audio-visual feature is 1536-d and here we adopt different N as 32, 64, 128, 256, 384, 512 and 768 together with a fixed $\gamma$ for each benchmark, and the AUROC value is reported as the evaluation metric. As depicted in Figure \ref{dimension}, the fitted curves reach the peak at $N=64$, $N=256$ and $N=256$ for VGGSound-GZSL$^{cls}$, UCF-GZSL$^{cls}$ and ActivityNet-GZSL$^{cls}$ benchmarks respectively and are generally ``flat'' which indicates the proposed EZ-OOD method is less sensitive with the dimension N. As a result, we can select this hyperparameter from a wide range of numbers with little influence on the final GZSL performance, which convincingly validates the excellent robustness of our method.

%% file: sec/5_conclusion.tex
\section{Conclusion}
\label{conclusion}
In this paper, we propose an extremely simple OOD detection based model EZ-AVOOD for Audio-Visual Generalized Zero-Shot Learning~(AV-GZSL) by ingeniously integrating the discriminative information held by class-specific logits and class-agnostic feature subspace. Superior experimental results on 3 audio-visual datasets fully demonstrate the effectiveness of our model. Moreover, the excellent compatibility of the proposed OOD detection method EZ-OOD is verified through deploying a different unseen classifier to construct a new model that outperforms the contrasting methods on both OOD detection performance and GZSL classification accuracy. Therefore, we conclude that EZ-AVOOD is new state-of-the-art of AV-GZSL.